# EMOTION DETECTION FROM TWEETS USING A BERT AND SVM ENSEMBLE MODEL

Ionuț-Alexandru ALBU[1], Stelian SPÎNU[2]

*Automatic identification of emotions expressed in Twitter data has a wide range of applications. We create a well-balanced dataset by adding a neutral class to a benchmark dataset consisting of four emotions: fear, sadness, joy, and anger. On this extended dataset, we investigate the use of Support Vector Machine (SVM) and Bidirectional Encoder Representations from Transformers (BERT) for emotion recognition. We propose a novel ensemble model by combining the two BERT and SVM models. Experiments show that the proposed model achieves a state-of-the-art accuracy of 0.91 on emotion recognition in tweets.*

## 1. Introduction

Emotion detection is a subfield of sentiment analysis and is concerned with detecting the writer's emotion from text data. The main difference between the two is that, whereas sentiment analysis tries to classify a text as being positive or negative, emotion detection tries to identify the exact emotion of the writer, such as happiness, sadness, fear, etc. Emotion analysis of tweets is a very challenging task for natural language processing systems. Unlike traditional text, tweets are very short messages and contain spelling mistakes, slang, shortened forms, phrasal abbreviations, and expressive lengthening [1]. Automatic identification of emotions expressed in Twitter data has a wide range of applications such as understanding consumer views regarding a product [2], stock market prediction [3], detection of depressive disorders [4], detecting bullying outbreaks [5], and identifying terrorist threats [6].

The contribution of this paper can be summarized as follows. We present an improved version of the WASSA dataset [7] for the task of emotion detection, that has been obtained through balancing and the addition of a new class, the neutral class. Our experimental results[3] show that SVM offers the best results when compared to other Machine Learning classifiers. Further, we fine-tune three

---

[1] The Military Technical Academy "Ferdinand I" of Bucharest, Romania, e-mail: alexalbu78@yahoo.com
[2] The Military Technical Academy "Ferdinand I" of Bucharest, Romania, e-mail: stelian.spinu@mta.ro
[3] https://github.com/alexalbu98/Emotion-Detection-From-Tweets-Using-BERT-and-SVM.git

BERT versions and present their performance on our dataset. Finally, we introduce a novel ensemble model which combines BERT and SVM models and achieves state-of-the-art results.

## 2. Related Work

Wang et al. [8] automatically created a large dataset containing about 2.5 million tweets, by leveraging the emotion hashtags. They applied two different classifiers, logistic regression and Naïve Bayes, to explore the effectiveness of various features such as n-grams, emotion lexicons, and part-of-speech information on the emotion identification task. The highest accuracy achieved was 0.6557.

Mohammad [9] automatically created a corpus of about 21,000 emotion-labelled tweets using hashtags. He employed binary SVMs, one for each of the six basic emotions of Ekman [10], and used the presence or absence of unigrams and bigrams as binary features. The binary classifiers were able to predict the emotions with a balanced F1-score of 0.499.

Janssens et al. [11] studied the impact of using weak labels compared to strong labels on emotion recognition for a corpus consisting of 341,931 tweets. The weak labels were created by employing the hashtags of the tweets and the strong labels by the use of crowdsourcing. The features extracted by combining n-grams and TF-IDF (Term Frequency-Inverse Document Frequency) were applied to five classification algorithms: Stochastic Gradient Descent, SVM, Naïve Bayes, Nearest Centroid, and Ridge. The results showed only a 9.25% decrease in F1-score when using weak labels.

Abdul-Mageed and Ungar [12] used distant supervision to automatically construct a large dataset of about 1.6 million labelled tweets and then trained Gated Recurrent Neural Networks for fine-grained emotion detection. They achieved an average accuracy of 0.8758 on 24 fine-grained types of emotions.

Felbo et al. [13] employed a variant of the Long Short-Term Memory model for emoji prediction on a dataset consisting of 1.6 billion tweets. Then they fine-tuned the pretrained model for emotion, sentiment, and sarcasm detection. Using their pretrained DeepMoji model, the highest averaged F1-score obtained in emotion analysis was 0.61.

Chiorrini et al. [14] investigated the use of Bidirectional Encoder Representations from Transformers (BERT) [15] for both sentiment analysis and emotion recognition of Twitter data. They achieved an accuracy of 0.90 on emotion recognition task.

### 3. Proposed Method
### 3.1 Dataset

The dataset used for training and evaluation of emotion in tweets was partly extracted from the WASSA dataset, which was offered to the participants in the Workshop on Computational Approaches to Subjectivity, Sentiment and Social Media Analysis (WASSA-2017) [7]. We extracted 1,500 tweets for each of the four emotions: fear, sadness, joy and anger, without taking into account data containing emotion intensities. Next to these four classes, we added an extra class of 1,500 neutral tweets, as it has been shown in [16] that having a neutral class in the classification process is important, because in this way the neural network will not have to classify unknown emotions into one of the learnt classes of emotions. Neutral tweets were extracted from CrowdFlower[4].

The reason we divided the dataset in this manner is the importance of having a well-balanced dataset. A well-balanced dataset contains an equal number of samples per class [17]. This ensures that the model will not favour larger classes in the classification process.

### 3.2 SVM Model

The preprocessing phase of the SVM model involves traditional machine learning preprocessing operations: emoticon to word conversion, Unicode to ASCII conversion, stop words filtering, sentence tokenization and vectorization, and label encoding. We first preprocessed tweets by eliminating unnecessary words and artefacts from tweets (usernames, links, hashtag symbols) and afterwards translating emoticons into their meaning by using Demoji Python library[5] for Unicode emoticons and meanings extracted from Wikipedia[6] for western-style emoticons. Emoticons to words conversion was used due to the fact that emoticons may be relevant in detecting the emotion in a tweet, as they are frequently used in text messages for better conveying feelings [18].

Unicode to ASCII conversion was deemed important as the ASCII character set is smaller compared to the Unicode one, therefore computation time is reduced, due to the fact that the word count will be reduced as well, which might result in a performance boost.

Stop words filtering in the dataset has been proven to improve performance and speed computation [19], so we used nltk's list of stopwords for this task. Stemming is another method of improving model performance by reducing derived words to a grammatical root called stem. In this article we used nltk's Snowball stemmer[7].

---

[4] https://data.world/crowdflower/sentiment-analysis-in-text
[5] https://pypi.org/project/demoji
[6] https://en.wikipedia.org/wiki/List_of_emoticons

For transforming each tweet into a numeric vector, we used tokenization alongside term-frequency times inverse document-frequency (tf-idf). *Tf-idf* [20] tries to estimate the importance of tokens in the dataset by computing two statistics: the *term frequency*, which means the appearance of a specific term in the tweet, and the *inverse document frequency* used to measure how much information that specific term provides relative to all the tweets in the corpus. The final result of the tf-idf is multiplication between the two frequencies.

In order to feed the data to the model, tweets, but also labels have to be preprocessed. Label encoding was achieved by using a numeric label encoder[8] for assigning to each emotion class a unique number.

For building the SVM model RBF (Radial Basis Function) kernel was chosen. The decision is based on the fact that its nonlinearity maps the data into a higher dimension space when compared to other kernels. It also has less hyperparameters and numerical difficulties than other kernels [21]. The regularization factor (C) was set empirically to 1.

### 3.3 BERT Model

In this paper three versions of BERT were analysed, vanilla BERT, which is the first version [15], RoBERTa (Robustly Optimised BERT approach) which is an optimised version of BERT [22], and BERTweet [23], which is a RoBERTa model pretrained on tweets.

For the BERT versions a different preprocessing approach was used. Links and usernames were eliminated, as those are considered unimportant and keeping them could result in a biased model. On top of this, BERT specific preprocessing was used, which includes:
- *Word tokenization*. Each sentence is split into its composing words and a unique word identifier is assigned. BERT uses a WordPiece tokenizer.
- *Padding*. Each sentence is padded to the same number of tokens.
- *Building the attention mask*. The attention mask will enable the model to distinguish between padding tokens and original tokens.
- *Adding BERT tokens*. Those tokens are CLS, SEP, PAD, UKN, EOS, and are part of how the model was pretrained using Masked Language Modelling [15].

Labels were encoded in the same manner as the SVM model, using a numeric label encoder.

For fine tuning the three BERT versions for the task of emotion detection, the base cased model was chosen for RoBERTa, BERTweet, and vanilla BERT.

---

[7] https://www.nltk.org/_modules/nltk/stem/snowball.html
[8] https://scikit-learn.org/stable/modules/generated/sklearn.preprocessing.LabelEncoder.html

Also, a pooled output was used for the BERT models utilizing the position of the first token, the [CLS] token. A pooled output is represented by a word embedding vector of size 768, which represents the word embedding of the [CLS] token. This is usually done when classifying the entire sequence, not individual tokens. On top of this output a dense layer was added, with an input dimension of 768, and an output dimension equal to the number of emotion classes. For getting the probabilities LogSoftmax[9] activation function was used as it offers better numeric stability than Softmax. During training a dropout layer with a dropout rate of 30% is added to mitigate overfitting. An overview of this fine-tuning approach can be seen in Fig. 1.

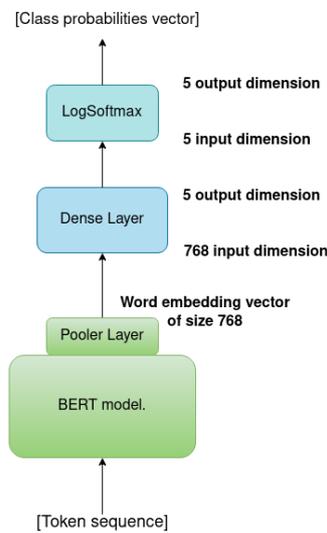

Fig. 1. The proposed BERT model

The chosen loss function was Negative Log Likelihood loss[10], as it can be used together with LogSoftmax to replace Categorical Cross Entropy[11]. The chosen optimizer was Adam with weight decay: learning rate was set to $2 \cdot 10^{-5}$ and a linear scheduler with no warmup steps was used, in order to change the learning rate dynamically as the training progresses. These configurations are in accordance with what has been shown in [15]. During training, gradient norm clipping with a max norm of 1 was used, to reduce the probability of appearance for the vanishing or exploding gradient.

### 3.4 BERT and SVM Ensemble Model

Ensemble models have been historically used to produce state of the art results for various machine learning classification problems [24]. Ensemble models offer better results as the advantages brought by each individual model are taken into account when producing an output. This usually works best when the composing models have different architectures, as this ensures that each model learns to look at different aspects of the training data [24].

---

[9] https://pytorch.org/docs/stable/generated/torch.nn.LogSoftmax.html
[10] https://pytorch.org/docs/stable/generated/torch.nn.NLLLoss.html
[11] https://pytorch.org/docs/stable/generated/torch.nn.CrossEntropyLoss.html

For building the proposed ensemble model, we combine a BERTweet model, as this version seems to offer the best results, and a SVM model. Each model will use its specific preprocessing before making predictions.

In order to obtain the probabilities for each emotion class, the ensemble model computes the sum of the log probability vectors obtained from the two models. Because log probabilities have values in the interval $(-\infty, 0]$, when adding the two probability vectors bad results will get worse and good results will not be as affected, due to the fact that those are closer to 0.

The workflow of the ensemble model can be observed in Fig. 2.

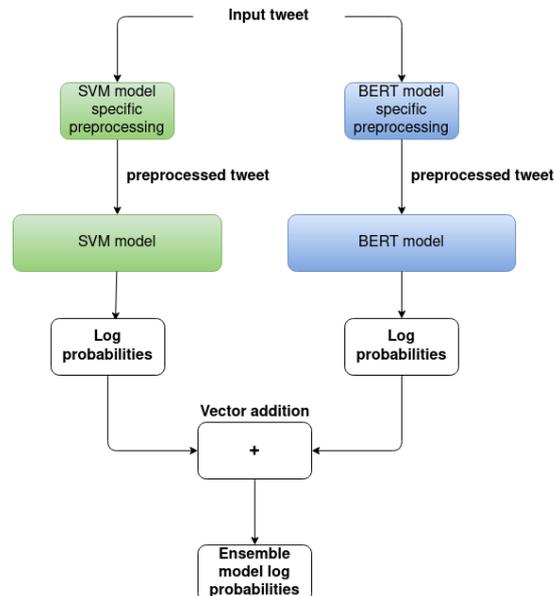

Fig. 2. Ensemble model workflow

### 4. Experimental Results

To gather insights upon our proposed model, first we check for the traditional machine learning model that has best performance on the dataset presented in Section 3. Afterwards we compare the three presented versions of BERT and finally we showcase the results of the ensemble model.

All models were trained using the same train-validation-test ratio of the data, which is: 1,200 samples for train, 150 for validation and 150 for testing, for each emotion.

### 4.1 Results for the SVM Model

From our results, the SVM classifier seems to offer best results for the task of emotion detection when compared to other machine learning classifiers such as Multinomial Naïve Bayes [25] and Gaussian Naïve Bayes [26], as seen in Table 1.

*Table 1*

**Comparison between machine learning classifiers**

| Model | Accuracy |
|---|---|
| Multinomial Naïve Bayes | 0.80 |
| Gaussian Naïve Bayes | 0.73 |
| SVM | 0.84 |

For training the Bayesian models, the same preprocessing technique was used as for the SVM model. For Multinomial Naïve Bayes an additive smoothing parameter of 1 was chosen, and for Gaussian Naïve Bayes a smoothing variable of 0.5. Those values were chosen empirically. Afterwards the models were fitted on the training set and tested. The results show that Multinomial Naïve Bayes is the better choice of the two models, with an accuracy of 0.80. During training the SVM model has been fitted on the training set after the preprocessing phase. On the validation set the model obtained an accuracy score of 0.87. In order to ensure that our model will perform well on previously unseen data, the model was tested on the test set. The accuracy score on the test set was 0.84, which shows that the model is reliable. In order to better understand how the model performs on detecting each emotion, a confusion matrix was computed on the test set, which can be seen in Fig. 3.

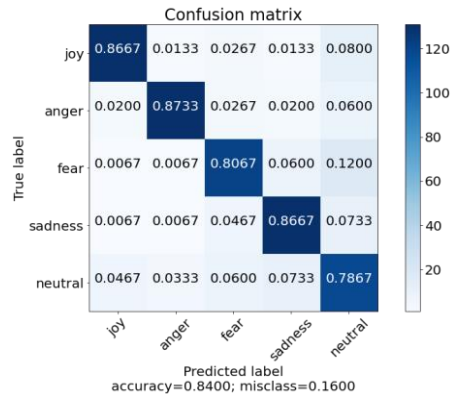

Fig. 3. Confusion matrix for the SVM model on the test set

*Table 2*

**Precision, recall, and F1-score for the SVM model on the test set**

|  | Joy | Anger | Fear | Sadness | Neutral |
|---|---|---|---|---|---|
| Precision | 0.92 | 0.94 | 0.83 | 0.84 | 0.70 |
| Recall | 0.87 | 0.87 | 0.81 | 0.87 | 0.79 |
| F1-score | 0.89 | 0.90 | 0.82 | 0.85 | 0.74 |

The results show that the model performs worst when classifying the neutral emotion and best when detecting anger. This can be due to the fact that the neutral class has a different token distribution than all the other classes. This behaviour has been previously noted for Multinomial Naïve Bayes classifier as well.

### 4.2 Results for the Proposed BERT Models

For training the proposed BERT models, a training time of 5 epochs and a batch size of 16 were used. The training time does not have to be long as the models will reach their maximum accuracy after relatively few epochs due to the fact that the models are pretrained.

For the *vanilla BERT* model, the tweets were padded during the preprocessing phase to a size of 85 tokens, which is the maximum size found in the presented dataset. For the validation set the model got an accuracy of 0.89, and for the testing set an accuracy of 0.87. This shows that a vanilla BERT model offers improved results compared to the SVM model.

For the *RoBERTa* model the tweets were padded during the preprocessing phase to a size of 170 tokens. This size increase is due to the fact that RoBERTa tokenizer will add extra spaces before emojis. The model got an accuracy of 0.87 for the testing and validation sets. This shows that a RoBERTa model does not bring any additional improvements when compared to the vanilla BERT model.

For the *BERTweet* model, the tweets were padded to a size of 90 tokens, which is the maximum size found in the presented dataset after the model applies its specific preprocessing. For the validation set the model got an accuracy of 0.89, and for the testing set an accuracy of 0.89. The confusion matrix on the test set can be seen in Fig. 4. This shows that BERTweet performs slightly better than the vanilla version. Therefore, this BERT version was chosen for building the BERT-SVM ensemble model. A comparison between BERT versions on the presented dataset can be observed in Table 4.

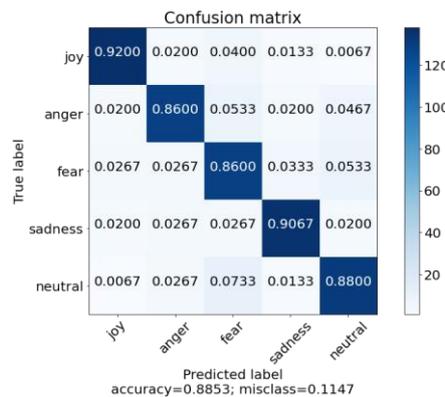

Fig. 4. Confusion matrix for the BERTweet model on the test set

*Table 3*
**Precision, recall, and F1-score for the BERTweet model on the test set**

|  | Joy | Anger | Fear | Sadness | Neutral |
|---|---|---|---|---|---|
| Precision | 0.93 | 0.90 | 0.82 | 0.92 | 0.87 |
| Recall | 0.92 | 0.86 | 0.86 | 0.91 | 0.88 |
| F1-score | 0.92 | 0.88 | 0.84 | 0.91 | 0.88 |

*Table 4*
**Comparison between BERT versions on the presented dataset**

| Version | Validation Accuracy | Test Accuracy |
|---|---|---|
| Vanilla BERT | 0.89 | 0.87 |
| RoBERTa | 0.87 | 0.87 |
| BERTweet | 0.89 | 0.89 |

### 4.3 Results for the Ensemble Model

For assessing the performance of the ensemble model, both the SVM and BERTweet models computed the log probability vectors for the validation set and the test set. Afterwards, the vectors of the two models were added in order to obtain the results of the ensemble model. The predicted emotion class for a tweet will be the one with the highest probability found in the corresponding probability vector. The ensemble model obtains an accuracy score of 0.91 for both validation and testing set. From the confusion matrix in Fig. 5, we can conclude that the model performs well on all classes, the lowest accuracy score being 0.84 for the neutral emotion class.

### 4.4 Comparison Between Emotion Detection Models

As it can be seen in Table 6, best results were obtained using the ensemble model. The model surpasses both SVM and BERT models with an accuracy of 0.91.

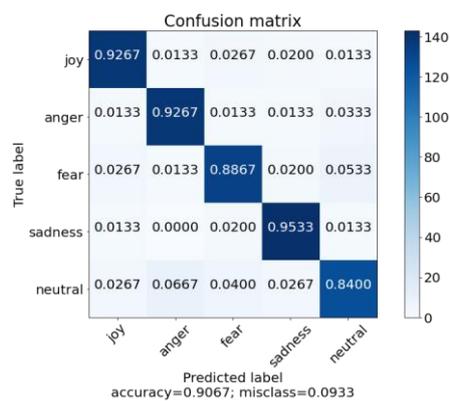

Fig. 5. Confusion matrix for the BERTweet-SVM model on the test set

This accuracy is high enough to consider the proposed model state-of-the-art. However, the accuracy is slightly lower when compared to the accuracy of the same model trained on the initial dataset, without additional neutral tweets, which is 0.94.

Table 5

**Precision, recall, and F1-score for the BERTweet-SVM model on the test set**

|  | Joy | Anger | Fear | Sadness | Neutral |
|---|---|---|---|---|---|
| Precision | 0.92 | 0.91 | 0.90 | 0.92 | 0.88 |
| Recall | 0.93 | 0.93 | 0.89 | 0.95 | 0.84 |
| F1-score | 0.92 | 0.92 | 0.89 | 0.94 | 0.86 |

Table 6

**Comparison between selected emotion detection models**

| Model | Accuracy |
|---|---|
| SVM | 0.84 |
| BERTweet | 0.89 |
| Proposed ensemble model | 0.91 |

Table 7

**Results obtained for the data set without a neutral class**

| Model | Accuracy |
|---|---|
| SVM | 0.91 |
| BERTweet | 0.90 |
| Proposed ensemble model | 0.94 |

It is worth mentioning that the proposed model trained on the extended dataset is able to detect when tweets do not convey any emotion, which might prove useful in real life usage of the model.

## 5. Conclusions

From our results, we can conclude that using an ensemble model for the task of emotion detection offers the best results when compared to other approaches researched so far. When compared to [14], our results on the same dataset are slightly better, with an accuracy score of 0.94.

The highest F1-score was obtained for the anger class in the case of the SVM classifier and for the joy class, respectively, in the case of the BERTweet model. The SVM and BERTweet model seem to improve on each other's results when combined, which is an important discovery when considering the future development of other ensemble models.

For future research, the ensemble model can be improved by researching BERT versions with more parameters such as the large version, which might lead to better results.